\documentclass[11pt]{article}
\usepackage[preprint]{acl}
\usepackage{times}
\usepackage{latexsym}
\usepackage{amsmath}
\usepackage{amsthm}
\usepackage{amsfonts}
\usepackage{comment}
\usepackage{booktabs}
\usepackage{float}
\usepackage{array}
\usepackage{multirow}
\usepackage[T1]{fontenc}
\usepackage{tabularx}
\usepackage[utf8]{inputenc}
\usepackage{microtype}
\usepackage{graphicx}

\title{From Error Prediction to Causal Influence: Auditing SAE Features in Multiple-Choice QA}

\author{
  \begin{tabular}{cc}
    \textbf{Aria Nourbakhsh} &
    \textbf{Adelaide Danilov} \\
    \textbf{Oleksandr Marchenko Breneur} &
    \textbf{Salima Lamsiyah}
  \end{tabular}
  \\[0.5em]
  Department of Computer Science (DCS), University of Luxembourg
  \\[0.4em]
  \textbf{Correspondence:}
  \href{mailto:aria.nourbakhsh@uni.lu}{aria.nourbakhsh@uni.lu}
}

\begin{document}
\maketitle

\begin{comment}
and model behavior can be measured by the logit margin between the correct answer and the strongest distractor    
\end{comment}

\begin{abstract}

Sparse autoencoders (SAEs) can predict internal features associated with model
errors, but association alone does not establish causal influence. We study
this distinction in multiple-choice question answering, where answer preference
can be measured by the logit margin between the correct answer and the
competing choices. Across eight datasets, we analyze Gemma~3 1B and 4B with
Gemma Scope SAEs. We propose a pipeline to filter out the features that have high correlation with format and content artifacts, such as prompt and answer permutation, and evaluate the remaining ones with causal feature interventions.
Our results show that SAE activations contain predictive signals about wrongness and
uncertainty across models and datasets. After filtering artifact-contaminated features,
selected features for causal intervention at question and option spans produce larger absolute margin
shifts and a higher wrong-to-correct rate than randomly selected features,
indicating greater causal leverage. However, answer flips are rare, and we do not observe stable effect directions
across meaning-preserving changes in prompt format and answer order. An adapted HalluSAE baseline produces
larger final-token margin changes, but relies on substantially contaminated
features and does not significantly improve correction rates. Therefore, the
identified features reflect context-dependent answer competition rather than the mechanism of incorrect answer errors.

\end{abstract}

\section{Introduction}

Large language Models (LLMs) can produce outputs that are incorrect or unsupported by the context~\cite{ji2023survey}. In multiple-choice evaluation, observed behavior can also change under semantically equivalent prompt forms or answer orders~\cite{pezeshkpour-hruschka-2024-large, ganesh-etal-2026-rethinking}. An incorrect answer may reflect a learned misconception~\cite{lin-etal-2022-truthfulqa}, uncertainty or competition between plausible candidates~\cite{kuhn2023semantic,farquhar2024detecting}, or a surface-form bias~\cite{yang-etal-2025-option}.

One of the tools to study the behavior of LLMs is Sparse autoencoders (SAEs). SAEs decompose language-model activations into sparse feature dictionaries~\cite{p86,p81,p87,shu-etal-2025-survey}. They support large-scale feature analysis and predictions
\cite{mcdougall2025gemmascope2,p90}, but predictive or attributive features need not be causal mechanisms~\cite{p84,p64,gulko-etal-2025-ce,p63}. A feature may correlate with an incorrect answer because it tracks uncertainty, answer format, option position, topic, or final-token decision pressure rather than because it causes the model to choose that answer~\cite{ICLR2024_6c0e99d7,ICLR2025_a712d461}. This distinction is the main idea of the current work.

In this paper, we explore three questions about SAE features: whether SAE activations predict errors,
whether intervening on a feature changes answer preferences, and whether
those changes correct errors. Forced-choice evaluation provides an
explicit output space for comparing answer
logits~\cite{stoehr-etal-2024-activation,wang-etal-2024-answer-c,ganesh-etal-2026-rethinking}. We explore these questions on two instruction-tuned Gemma~3 models~\cite{gemma_team_2025_gemma3}; we combine
behavioral probes, artifact-risk screening, contrastive direct logit
attribution, and instance interventions to answer the above questions.

Our results show that SAE features can predict forced-choice errors, but many of them are entangled with format, topic, or answer-choice structure. Only question- and option-span interventions produce larger distributional shifts, and only a small number of samples flip to the correct answer.

Hence, our contributions are:
\begin{itemize}
    \item We introduce a pipeline for discovering behaviorally
    predictive SAE features across eight datasets and six token positions. This pipeline combines multiple stages to select predictive and causal features for the model's answer choices while filtering out features that are highly correlated and contaminated by the datasets, prompt, answer letter, and other artifacts. %Your wording is good. Please put it here

    \item We causally evaluate the filtered features using steering. We found that features extracted from question and option
    spans produce larger margin changes and a higher
    wrong-to-correct rate than random-active controls, but answer flips are rare, and effect directions are unstable across steering methods, prompt templates, and answer permutations.

    \item We quantify artifact entanglement in both our candidates and an adapted HalluSAE~\cite{p90} baseline. HalluSAE is predictive and produces large final-token margin shifts; however, $95.5\%$ of its nonzero feature units fail our artifact-risk criterion. This shows that predictive performance and logit influence do not by
    themselves establish artifact-independent causal mechanisms.
\end{itemize}

\section{Related Work}

\paragraph{Sparse Autoencoders for Language Model Interpretability. }

Sparse autoencoders are among the promising methods for decomposing dense language-model activations into interpretable latent features. Early work showed that SAEs can recover features that are more interpretable than individual neurons~\cite{p86}, while later work improved reconstruction and sparsity tradeoffs through larger-scale training and architectural variants such as Gated SAEs and JumpReLU SAEs~\cite{p81,p87,p53}. 

At a high level, an SAE is trained on internal model activations to reconstruct a dense activation vector from a much larger but sparse set of latent features. Given a model activation $x \in \mathbb{R}^{d}$, the encoder maps it to nonnegative feature activations $z \in \mathbb{R}^{m}$, often with $m \gg d$:
\begin{gather}
    z = \sigma\!\left(W_{\mathrm{enc}}(x - b_{\mathrm{dec}}) + b_{\mathrm{enc}}\right), \label{eq:sae-encoder} \\ 
    \hat{x} = b_{\mathrm{dec}} + W_{\mathrm{dec}} z. \label{eq:sae-decoder}
\end{gather}
The SAE is trained to reconstruct $x$ while keeping $z$ sparse, for example with an $L_{1}$-regularized objective:
\begin{equation} \label{eq:sae-objective}
    \mathcal{L}_{\mathrm{SAE}} = \|x - \hat{x}\|_{2}^{2} + \lambda \|z\|_{1}.
\end{equation}
Here, each column of $W_{\mathrm{dec}}$ can be interpreted as a learned feature direction, and the scalar $z_i$ indicates how strongly feature $i$ is active on a given token. 

Several recent methods uses SAE features for steering, including activation additions, refusal steering, in-context-learning improvement, and knowledge-selection control~\cite{p93,p99,obrien2025steering,cho-hockenmaier-2025-toward,zhao-etal-2025-steering}. These studies show that sparse directions can sometimes be used as actionable control handles.

\paragraph{Mechanistic analyses of multiple-choice answering.}
Prior work distinguishes answering a question from producing its associated
option label. \citet{lieberum2023doescircuitanalysisinterpretability} identify attention heads that promote
the correct answer letter.
\citet{wiegreffe2025answer} use activation patching and vocabulary projections
to distinguish answer-symbol selection from its subsequent promotion in the
output distribution. \citet{wong-etal-2026-models} identify linearly decodable
option-correctness signals and evidence that selecting answer content precedes
mapping it to an output symbol. Related causal analyses identify attention
heads contributing to option-symbol bias~\citep{yang-etal-2025-option}.
These findings motivate distinguishing knowledge-related behavior from
format-dependent processing. Our study complements this work by testing
whether behaviorally predictive SAE features, screened for potential
artifacts, support reliable error correction under feature interventions
and changes in prompt format.

\paragraph{Evaluation, Artifacts, and Controls.}
The SAE literature increasingly stresses that feature discovery pipelines need rigorous controls. SAEBench and related evaluation work argue for task-specific benchmarks rather than relying only on qualitative feature labels~\cite{p64,p84,gulko-etal-2025-ce}. Sparse probing work questions when sparse features are actually useful for known downstream concepts~\cite{p63,p101}, and work on feature explanation falsification highlights the risk that a single description can overstate what a feature reliably represents~\cite{p89}.

Our work builds on this line of work by asking a narrower causal question: whether features associated with wrong forced-choice answers affect logit difference and preference between correct vs. wrong answers. Because predictive features can still be non-causal, polysemantic, or artifact-driven~\cite{fereidouni-etal-2026-evaluating,p63}, we use ablation and steering as diagnostic tests and incorporate artifact controls before interpreting a feature mechanistically.

\section{Methodology}
\label{sec:methodology}

\begin{figure*}[t]
    \centering
    \includegraphics[width=\textwidth]{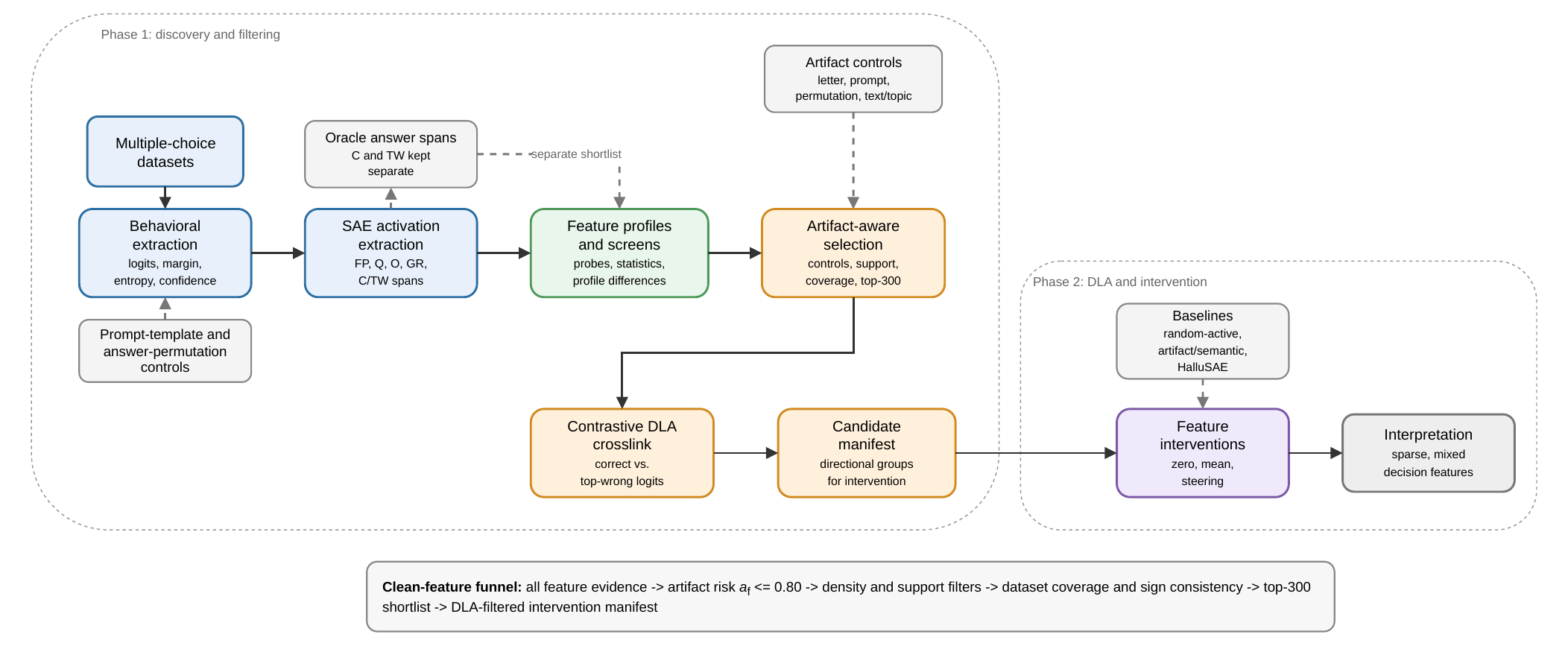}
    \caption{Overview of the discovery and intervention pipeline. Behavioral
    measurements and SAE activations feed predictive, statistical, and
    artifact-aware screening. Contrastive DLA then prioritizes separate
    non-oracle and oracle candidates for feature interventions and baseline
    comparisons. Position codes are FP (final prompt token), Q (question
    span), O (option span), GR (generated reasoning), C (correct-choice text),
    and TW (top-wrong-choice text); C and TW are oracle spans.}
    \label{fig:methodology_pipeline}
\end{figure*}

This work has two main phases. Phase~1 is used for feature candidate discovery and selection. In Phase~2, we perform causal interventions. Figure~\ref{fig:methodology_pipeline} summarizes the full pipeline.

\subsection{Phase 1} \label{sec:methodology_phase_1}

\paragraph{Behavioral extraction.}
For each multiple-choice prompt $i$, let $\mathcal{L}_i$ be the available answer letters and $y_i$ the gold letter. Let $g_i(v)$ denote the next-token logit for token $v$ at the answer position. For each answer letter $\ell\in\mathcal{L}_i$, let $T(\ell)$ be its set of valid single-token encodings. We pool their logits and normalize across the available letters:

\begin{gather}
    z_{i,\ell} = \log\sum_{v\in T(\ell)} e^{g_i(v)},\label{eq:answer_letter_logit} \\
    p_i(\ell) = \frac{e^{z_{i,\ell}}}{\sum_{k\in\mathcal{L}_i}e^{z_{i,k}}}.
\end{gather}
Let $w_i$ be the highest-scoring non-gold letter. The quantities
$z_{i,y_i}$ and $z_{i,w_i}$ are the pooled logits from
Eq.~\ref{eq:answer_letter_logit}, evaluated at the gold letter $y_i$ and
the strongest wrong letter $w_i$, respectively. Their difference defines the correct-vs.-top-wrong margin:
\begin{equation}
    m_i=z_{i,y_i}-z_{i,w_i}.
\end{equation}
We additionally record $p_i(y_i)$, $p_i(w_i)$, and answer-letter entropy $H_i=-\sum_{\ell\in\mathcal{L}_i} p_i(\ell)\log p_i(\ell)$.

\paragraph{SAE feature extraction.}

We encode residual-stream states using pretrained Gemma Scope 2
SAEs~\cite{mcdougall2025gemmascope2}. Each behavioral instance \(i\)
corresponds to one rendered prompt under a particular template, answer
permutation, and response mode (Sec.~\ref{sec:prompting_positions}). For
feature \(f\) and token set \(P_i\), we max-pool token activations as a feature may occur multiple times in different spans (Sec.~\ref{sec:prompting_positions}):
\begin{equation}
    A_{i,f}=\max_{t\in P_i}a_{i,t,f}.
\end{equation}
where \(a_{i,t,f}\) is the SAE activation of feature \(f\) at token
position \(t\) in behavioral instance \(i\).

\paragraph{Feature discovery.}
Within each dataset--model--SAE--response-mode--position setting, we summarize each feature by its activation support and frequency, mean and maximum activation, class-conditional means, and correlations with $m_i$, $H_i$, $p_i(y_i)$, $p_i(w_i)$, $z_{i,y_i}$, and $z_{i,w_i}$.

We use two complementary analyses. The \emph{candidate-ranking probe}
fits $L_1$-regularized logistic regressions for four binary targets: wrong
vs.\ correct (WvC), low vs.\ high margin (LMvHM), wrong high-confidence vs.\
correct high-confidence or high-margin (WHC), and wrong vs.\ correct among
low-margin instances (WLM). Ridge regressions are used for the continuous
margin and entropy targets. A separate \emph{statistical audit} fits the
same six targets using at most 1{,}000 features per setting, prioritizing
features active on the largest number of behavioral instances. We also compute per-feature Pearson and Spearman correlations to measure linear and monotonic associations with each behavioral target, bootstrap
intervals for binary mean-activation differences, and mutual-information
estimates. Appendix Table~\ref{tab:stat_screen_target_definitions} gives the
exact target definitions and thresholds.

\paragraph{Artifact controls.}
Multiple-choice prompts can create format-related shortcuts, so we evaluate four possible confounds for each feature active on at least 20 behavioral instances. Each control produces a raw score, which we convert to a within-run percentile.

The option-letter score \(q_f^{\mathrm{letter}}\) measures whether activation
depends on the correct answer label after subtracting the mean activation
within matched example--template groups. For example, a feature that
consistently activates more when the correct answer is labeled A receives a
high option-letter score. The prompt score \(q_f^{\mathrm{prompt}}\) measures
dependence on the prompt template after the corresponding adjustment within
matched example--permutation groups. The permutation score
\(q_f^{\mathrm{perm}}\) measures instability across answer-order permutations,
and the text score \(q_f^{\mathrm{text}}\) is the strongest association with
question categories, text lengths, answer-text similarities, or TF-IDF/SVD
text components.

We summarize the four controls with
\begin{equation}
    a_f=\max\{q_f^{\mathrm{letter}},q_f^{\mathrm{prompt}},
    q_f^{\mathrm{perm}},q_f^{\mathrm{text}}\}.
\end{equation}
For example, \(q_f^{\mathrm{letter}}=.90\) means that the feature has stronger option-letter dependence than \(90\%\) among the tested features in that setting.  We take the maximum score as strong evidence that any single confound is sufficient to assign a high risk. 
Appendix~\ref{app:artifact_score} gives the exact control statistics,
covariates, and clean-feature eligibility criteria.

\paragraph{Predictive shortlist.}
For each model--layer--SAE-width--feature unit \(f\), we aggregate
validation-weighted candidate-ranking evidence across datasets, response modes, and
token positions. A unit is eligible only if its maximum artifact risk is at
most \(0.80\), its activation frequency is below \(0.95\) in every contributing
setting, it is active on at least 100 behavioral instances in one setting, and
it has evidence from at least two datasets. We also require coefficient-sign
consistency of at least \(0.60\). Within each token position, this consistency
is
\begin{equation}
\label{eq:sign_consistency}
c_f=\frac{\left|\sum_r s_{f,r}\right|}{\sum_r|s_{f,r}|}.
\end{equation}
where \(s_{f,r}\) is signed, validation-weighted evidence derived from a candidate-ranking probe
coefficient; the unit-level value is averaged across contributing positions.
We rank eligible units by artifact-penalized evidence and retain the top 300
separately for the non-oracle and oracle pools.

\paragraph{DLA-based selection.}
After artifact-aware top-$k$ selection, we use contrastive direct logit attribution (DLA) as another heuristic before interventions. DLA specifies whether an active feature's decoder direction points toward the correct-answer logit or the strongest wrong-answer logit. A feature may be more active on incorrect instances while its decoder direction has little direct alignment with the correct-versus-wrong logit contrast; it can therefore predict errors without directly shifting the model toward either answer. Let $d_f$ be the decoder direction of feature $f$, and $u_\ell$ the mean unembedding direction for the valid token variants of answer letter $\ell$. DLA is defined as:
\begin{equation}
    C^{\mathrm{DLA}}_{i,f} = A_{i,f}\,d_f^\top(u_{y_i}-u_{w_i}).
\end{equation}
Positive values support the correct-vs. top-wrong margin, negative values support the top-wrong answer, and magnitude measures alignment strength. We then retain features satisfying fixed directional-consistency or high-magnitude mixed-direction criteria, detailed in
Appendix~\ref{app:dla_thresholds}.

\subsection{Phase 2}
\label{sec:methodology_phase_2}

\paragraph{Feature interventions.}
For a selected feature $f$ and intervention span $P_i$, we target a nominal activation $\tau_{i,t,f}$ by applying the decoder-vector update
\begin{equation} \label{eq:feature_intervention}
    r'_{i,t} = r_{i,t}+(\tau_{i,t,f}-a_{i,t,f})d_f, \quad t\in P_i.
\end{equation}
Here, $r_{i,t}$ is the residual-stream vector at token position $t$ in
instance $i$ at the intervention layer, and $r'_{i,t}$ is its updated value.

We test zero replacement (\(\tau_{i,t,f}=0\)), mean replacement
(\(\tau_{i,t,f}=\mu_f\)), and upward steering
(\(\tau_{i,t,f}=a_{i,t,f}+\lambda\sigma_f\)). Here, \(\mu_f\) and
\(\sigma_f\) are the mean and standard deviation of the row-level max-pooled
activation \(A_{i,f}\) across behavioral instances in the same
dataset--model--layer--SAE-width--response-mode--position configuration.

\section{Experiments}
\label{sec:experimental_setup}

\subsection{Datasets and splits}
\label{sec:datasets}

We evaluate on eight multiple-choice datasets, summarized in
Table~\ref{tab:datasets}. We select 4000 random samples, cap them at most 2000 characters, and convert each sample to the same lettered-choice interface.

\begin{table}[t]
\centering
\scriptsize
\setlength{\tabcolsep}{3pt}
\resizebox{\columnwidth}{!}{%
\begin{tabular}{llrc}
    \toprule
    Dataset & Source split & $N_{\mathrm{samples}}$ & Choices \\
    \midrule
    CommonSenseQA     & train       & 4{,}000 & 5 \\
    TruthfulQA MC1    & validation  & 817         & 2--13 \\
    HaluEval QA       & data        & 4{,}000 & 2 \\
    Wiki-FACTOR       & data        & 1{,}560 & 4 \\
    Med-HALT R-NOTA   & dev         & 2{,}636 & 4 \\
    MMLU              & validation  & 1{,}531 & 4 \\
    SciQ              & train       & 4{,}000 & 4 \\
    FEVER             & dev         & 4{,}000 & 3 \\
    \bottomrule
\end{tabular}%
}
\caption{Datasets, source splits, sample counts, and numbers of answer choices
before prompt expansion.}
\label{tab:datasets}
\end{table}

CommonSenseQA tests commonsense reasoning among plausible alternatives
\cite{talmor2019commonsenseqa}. TruthfulQA MC1 tests whether a model favors true answers over common misconceptions \cite{lin-etal-2022-truthfulqa}. MMLU covers academic and professional knowledge across different subjects \cite{hendrycks2021measuring}. SciQ contains science questions; we use only its questions and answer choices without the supporting passage
\cite{welbl-etal-2017-crowdsourcing}. HaluEval QA asks the model to distinguish a supported answer from a
hallucinated one using the supplied passage \cite{li2023halueval}. Wiki-FACTOR
contrasts factual Wikipedia continuations with contradictory alternatives
\cite{muhlgay2023factor}. Med-HALT R-NOTA evaluates errors in medical reasoning
\cite{pal2023medhalt}. FEVER is converted to classification among
\emph{Supported}, \emph{Refuted}, and \emph{Not enough information}, without
evidence sentences \cite{thorne-etal-2018-fever}. 

We split original examples into discovery, validation, and test partitions
(approximately 60/20/20).

\subsection{Prompting and activation positions}
\label{sec:prompting_positions}

Different prompts can change the behavior of the models~\cite {ICLR2024_6c0e99d7}. Therefore,
We use three prompt templates that vary option-label syntax and the answer cue. Options are formatted as \texttt{A.}, \texttt{(A)}, or \texttt{A:}, accompanied by cues such as \texttt{Answer:}, \texttt{The correct option is}, or \texttt{Final answer:}. Their structures and response cues are given in Appendix Table~\ref{tab:prompt_templates_exact}. 

As LLMs are sensitive to the positions of answer letters \cite{pezeshkpour-hruschka-2024-large, molfese-etal-2025-right}, we apply 3 permutations to the answer letters and their order, adding to the original order of the answers in the dataset.

We also evaluate three response modes: In \emph{forced letter} (FL), the prediction is the highest-scoring answer letter at the next-token position. In \emph{reason first} (RF), the model generates a rationale followed by a final letter. In \emph{letter then reason} (LTR), it generates the letter before its explanation. Both generative modes use greedy decoding with at most 80 new tokens. Generated answers are parsed to response letters; failed RF parses are recovered using an appended \texttt{Final answer:} cue. For all modes, margin, answer probabilities, and entropy are computed from the answer-letter logits.  Table~\ref{tab:activation_positions} summarizes the six activation positions we extract the SAE features.

After sampling and length filtering, we extract SAE activations from 1{,}472{,}694. rendered prompt instances.

\begin{table}[t]
\centering
\small
\setlength{\tabcolsep}{3pt}
\begin{tabular}{@{}lll@{}}
    \toprule
    Code & Token position & Role \\
    \midrule
    FP & final prompt token     & primary decision readout \\
    Q  & question/context span  & non-oracle context \\
    O  & complete option block  & non-oracle choices \\
    GR & generated rationale    & non-oracle generation \\
    C  & gold-choice text       & oracle diagnostic \\
    TW & top-wrong-choice text  & oracle diagnostic \\
    \bottomrule
\end{tabular}
\caption{Reported activation positions. Multi-token feature activations are
aggregated by taking their maximum over the corresponding span.}
\label{tab:activation_positions}
\end{table}

\subsection{Models and SAEs}
\label{sec:models_saes}

We study the instruction-tuned Gemma 3 models at 1B and 4B scales~\cite{gemma_team_2025_gemma3}. For each model, we use the corresponding Gemma Scope 2 residual-stream SAEs~\cite{mcdougall2025gemmascope2}. We analyze layers 17 and 22 for Gemma 3 1B and layers 22 and 29 for Gemma 3 4B, using 16K- and 65K-width SAEs at each layer.

\subsection{Feature Selection}
\label{sec:feature_selection_setup}

We construct separate non-oracle (FP, Q, O, and GR) and oracle (C and TW)
candidate pools. For each dataset, model, layer, SAE width, response mode, and
position, we fit the six candidate-ranking probes from
Section~\ref{sec:methodology_phase_1} on discovery instances, using fixed
regularization ($C=.1$ for classification and $\alpha=100$ for regression).
From each candidate-ranking probe, we retain up to 100 features ranked by
absolute coefficient. A feature at rank \(r\) contributes
\(Q/\log_2(r+1)\), where \(Q\) is the candidate-ranking probe's validation
quality (Appendix~\ref{app:feature_ranking}). We then group records
for the same model--layer--SAE-width--feature unit across targets, datasets,
response modes, and positions. The aggregate score rewards repeated,
directionally consistent evidence and penalizes artifact risk, dense
activation, and discovery--validation gaps. 

We apply the eligibility criteria described in
Section~\ref{sec:methodology_phase_1} to form separate non-oracle and
oracle shortlists. DLA selection uses only discovery and validation
instances, with the thresholds specified in
Appendix~\ref{app:dla_thresholds}, yielding 55 non-oracle and 22 oracle
feature units. Interventions are evaluated only on test instances.

\subsection{Intervention Setup}
\label{sec:intervention_setup}

We intervene on one SAE feature at a time in forced-letter runs using the
residual update in Eq.~\ref{eq:feature_intervention}. For each
dataset--model--layer--SAE-width--feature--position combination, we max-pool its
saved token activations within each behavioral instance and select up to ten
active test instances with the largest pooled values. The edit is applied to
every token in the corresponding span of FP, Q, or O for non-oracle experiments,
and C or TW for oracle spans.\footnote{GR contributes to feature
discovery and DLA selection but is not an intervention span. In
letter-then-reason mode it occurs after the answer; in reason-first mode,
causal testing would require online intervention during autoregressive
generation, which the present fixed-span implementation does not support.}

For each selected instance, we compare zero replacement (\(\tau=0\)), mean
replacement (\(\tau=\mu_f\))~\cite{marks2025sparse}, and upward steering
(\(\tau=a_{i,t,f}+\lambda\sigma_f\))~\cite{obrien2025steering} for
\(\lambda\in\{1,2,4\}\) against a no-intervention pass on the same prompt.
For any outcome \(x\), we define
\(\Delta x=x^{\mathrm{int}}-x^{\mathrm{base}}\) intervention minus no-intervention. We denote the change
in the correct-vs.-strongest-wrong margin by \(\Delta m\), changes in the gold
and strongest-wrong logits by \(\Delta z_y\) and \(\Delta z_w\), changes in
their probabilities by \(\Delta p_y\) and \(\Delta p_w\), and the change in
answer-letter entropy by \(\Delta H\). A positive \(\Delta m\) indicates a
shift toward the correct answer.

\subsection{Baselines}
\label{sec:baselines}

We use two baselines. Random-active intervention controls sample five eligible features that are active on the test split for each forced-letter FP, Q, or O run. For each feature, we select up to ten test instances with the highest activation and apply the same interventions used for selected features. Also, for comparison, we adapted the HalluSAE baseline~\cite{p90}, which computes DLA on discovery instances for each dataset--model--SAE-width setting at FP and selects 100 features in total across the two selected layers. It then fits a class-balanced \(L_1\) logistic wrong-vs.-correct classifier, selecting its regularization by five-fold cross-validation on discovery data and reporting validation and test performance. We use layers 17/22 for 1B and 22/29 for 4B instead of reproducing HalluSAE's Stage-I layer localization. We additionally run ten random-active selection repetitions, each containing 100 features. For the causal FP comparison, we match the number of HalluSAE and selected run-specific feature tests within each dataset--model--layer--SAE-width  and apply the same interventions on test instances.

\section{Results and Discussion}
\label{sec:results}

\subsection{MCQA Performance}
The performance of the two Gemma models on each dataset differs substantially. We define the
wrong-high-confidence (WHC) rate as the percentage of all test instances on
which the model answers incorrectly while assigning at least \(0.70\)
probability to its strongest wrong answer. Macro-averaged across datasets,
accuracy is \(43.8\%\) for the 1B model and \(58.6\%\) for the 4B model; the
corresponding WHC rates are \(52.0\%\) and \(39.3\%\). The 4B model has higher
accuracy and a lower WHC rate on every dataset. Appendix
Table~\ref{tab:forced_letter_behavior} reports the complete dataset--model
breakdown.

\subsection{Statistical Audit}
\label{sec:statistical_discovery_results}

We report the statistical audit described in
Section~\ref{sec:methodology_phase_1}, using up to 1{,}000 features per setting
and evaluating discovery-trained models on validation and test instances.

Table~\ref{tab:stat_screen_main_results} reports final-prompt-token (FP) test
performance over 192 runs. LMvHM and WHC have the highest mean AUROC (.739),
followed by WvC (.715) and WLM (.706). WLM has a higher F1 (.805) despite its
lower AUROC, consistent with the class imbalance induced by restricting the
task to low-margin instances. Entropy is more predictable than margin: its
median $R^2$ is .285, compared with .117 for margin.

\begin{table}[H]
\centering
\small
\setlength{\tabcolsep}{5pt}
\begin{tabular}{lcc}
    \toprule
    Target & Metric & Test summary \\
    \midrule
    WvC    & AUROC / F1 & .715 / .644 \\
    LMvHM  & AUROC / F1 & .739 / .711 \\
    WHC    & AUROC / F1 & .739 / .688 \\
    WLM    & AUROC / F1 & .706 / .805 \\
    Margin & $R^2$ / $r$ & .117 / .357 \\
    Entropy & $R^2$ / $r$ & .285 / .541 \\
    \bottomrule
\end{tabular}
\caption{FP test performance over 192 runs: eight datasets, eight
model--layer--SAE configurations, and three response modes. Binary metrics are
means; regression $R^2$ and Pearson $r$ are medians.}
\label{tab:stat_screen_main_results}
\end{table}

Appendix~\ref{app:statistical_screen} gives the exact statistical-audit setup,
and Table~\ref{tab:stat_screen_target_definitions} defines the six targets.
Tables~\ref{tab:stat_screen_position_binary}
and~\ref{tab:stat_screen_position_regression} report all six activation
positions, and Tables~\ref{tab:stat_screen_dataset_position_binary}
and~\ref{tab:stat_screen_dataset_position_regression} provide the full
eight-dataset breakdown.

\subsection{Feature Selection}
\label{sec:feature_discovery_results}

Table~\ref{tab:feature_discovery_top300} summarizes the candidate-ranking
aggregation described in Section~\ref{sec:feature_selection_setup}.

Aggregation produced 81{,}304 non-oracle and 48{,}888 oracle units. After the
clean-candidate filters in Sec.~\ref{sec:feature_selection_setup}, 2{,}205
non-oracle and 1{,}100 oracle units remained, from which we retained the top
300 in each pool. Their mean artifact risks are .712 and .724, respectively.
Appendix
Figure~\ref{fig:artifact_score_vs_evidence} shows that strong unpenalized
predictive evidence also occurs above the artifact-risk gate, motivating the
separate artifact filter.

Among the non-oracle top 300, the strongest evidence comes from FP for 99
units, O for 93, Q for 56, and GR for 52. The oracle shortlist is nearly
151 for C and 149 for TW 149.

\begin{table}[H]
\centering
\resizebox{\columnwidth}{!}{
\begin{tabular}{lrrrr}
    \toprule
    Pool & Ranked & Eligible & Selected & Mean $a_f$ \\
    \midrule
    Non-oracle & 81{,}304 & 2{,}205 & 300 & .712 \\
    Oracle C/TW & 48{,}888 & 1{,}100 & 300 & .724 \\
    \bottomrule
\end{tabular}
}
\caption{Validation-ranked feature selection over eight datasets. Ranked
counts are aggregated feature units; eligible units pass all clean-candidate
filters. $a_f$ is artifact risk.}
\label{tab:feature_discovery_top300}
\end{table}

Finally, we compute DLA on discovery and validation instances for all 600 shortlisted features. Applying the fixed direction, magnitude, and rank criteria in Appendix~\ref{app:dla_thresholds} retains features with consistent correct- or top-wrong-answer alignment, together with features having large but mixed-direction attribution. This yields 55 distinct non-oracle and 22 distinct oracle feature units for intervention.

\subsection{Causal Interventions}
\label{sec:causal_intervention_results}

Single-feature interventions cause measurable but sparse changes whose
directions are mixed. Table~\ref{tab:phase2_main_interventions} places the non-oracle,
random-active, and oracle results in one comparison. W2C, C2W, and W2W denote
wrong-to-correct, correct-to-wrong, and wrong-to-different-wrong flips,
respectively.

\begin{table*}[t!]
\centering
\small
\setlength{\tabcolsep}{4.5pt}
\begin{tabular}{llrrrrrrrr}
    \toprule
    Set & Pos. & $N$ & $U$ & W2C & C2W & W2W
    & $\overline{\Delta m}$ & $\overline{|\Delta m|}$
    & $\overline{|\Delta p_y|}$ \\
    \midrule
    Selected & FP & 2{,}500  & 73  & 2  & 9  & 9  & +.016 & .154 & .0031 \\
    Random   & FP & 9{,}950  & 320 & 18 & 22 & 19 & +.007 & .237 & .0039 \\
    HalluSAE & FP & 3{,}650  & 73  & 5  & 9  & 3  & -.275 & 1.760 & .0052 \\
    \midrule
    Selected & Q  & 21{,}110 & 425 & 16 & 17 & 8  & +.002 & .158 & .0010 \\
    Random   & Q  & 15{,}150 & 320 & 1  & 9  & 10 & -.001 & .135 & .0011 \\
    Selected & O  & 20{,}170 & 415 & 19 & 19 & 23 & -.005 & .216 & .0021 \\
    Random   & O  & 13{,}395 & 320 & 7  & 4  & 3  & -.004 & .159 & .0016 \\
    \midrule
    Oracle   & C  & 6{,}805  & 145 & 6  & 9  & 4  & -.003 & .132 & .0012 \\
    Oracle   & TW & 7{,}790  & 160 & 0  & 5  & 0  & -.002 & .135 & .0009 \\
    \bottomrule
\end{tabular}
\caption{Final test-set intervention outcomes. $N$ counts
feature--instance--intervention events and $U$ counts run-specific feature
tests; each feature--instance pair contributes five intervention events.
$p_y$ is the gold-letter probability. HalluSAE matches the selected FP
candidate count within dataset--model--layer--width strata, but not the number
of active instances. Oracle C/TW spans have no random-active counterpart.}
\label{tab:phase2_main_interventions}
\end{table*}

\paragraph{Selected versus random.}
Table~\ref{tab:phase2_main_interventions} reports the position-specific event
counts and average shifts. Appendix Table~\ref{tab:phase2_uncertainty}
reports the pooled selected--random differences and their bootstrap intervals.
Pooling interventions at Q and O, selected features produce a larger mean
absolute margin change than random-active controls ($.187$ vs. $.146$). They
also correct a larger fraction of initially wrong intervention events
($.189\%$ vs. $.058\%$). Across 5{,}000 bootstrap resamples of
model--layer--width--feature units, the selected-minus-random differences are
$+.041$ $[.014,.067]$ for mean $|\Delta m|$ and $+.132$ percentage points
$[+.024,+.257]$ for the W2C rate. Both intervals lie above zero, providing
evidence that selected features exceed random controls on these two outcomes.
By contrast, the intervals for the C2W rate and $|\Delta p_y|$ include zero,
so these outcomes show no clear selected--random difference. At FP, selected
features produce a smaller mean absolute margin change than random features
($.154$ vs. $.237$; difference $-.083$ $[-.118,-.047]$). Thus, selected
features outperform random controls specifically on Q/O mean $|\Delta m|$ and
W2C rate; this advantage does not extend to all outcomes or to FP. The matched
HalluSAE comparison is reported in Appendix
Table~\ref{tab:hallusae_causal_fp}.

\paragraph{Effects across steering strengths, positions, and surface forms.}
Appendix Figure~\ref{fig:phase2_dose_response} reports the steering-strength
results, and Appendix Table~\ref{tab:phase2_pair_level} reports outcomes at
both the event and feature--prompt-pair levels. Each pair consists of one SAE
feature applied to one test prompt and evaluated under five intervention
variants.
Selected Q/O $|\Delta m|$ increases from $.164$ at $+1\sigma_f$ to $.268$ at
$+4\sigma_f$, but only $51.8\%$ of pairs increase monotonically. After
adjustment for intervention size and context, the selected--random
margin-change difference remains distinguishable at O ($.049$ $[.014,.084]$),
but not at Q or pooled Q/O (Table~\ref{tab:phase2_adjusted_effects}).
Selected--random differences in sign agreement across matched permutations
and templates also have intervals including zero. Oracle C/TW interventions
include both repairs and damage (6 W2C and 14 C2W events), without a matched
random-active comparison. These results support local influence on answer
preferences, but do not establish a general correction advantage.

\subsection{HalluSAE Baseline Comparison}
\label{sec:hallusae_baseline_results}

Appendix Table~\ref{tab:hallusae_predictive_artifact_audit} summarizes the
baseline's predictive performance, artifact audit, and overlap with our
selected features.
Most features retained by the adapted HalluSAE classifier do not pass our
artifact-risk criterion. Of its 711 nonzero-coefficient
model--layer--SAE-width--feature units, 679 (\(95.5\%\)) exceed
\(a_f=0.80\) in every forced-letter FP setting in which they are selected.
Only nine units remain below this threshold in every selected setting. The
HalluSAE units also have zero overlap with either our top-300 shortlist. Thus, HalluSAE's predictive features are largely
distinct from those retained by our artifact-aware selection.

Despite this difference, HalluSAE is predictive: across 32
dataset--model--width groups, it obtains mean test AUROC/F1 of
\(.775/.674\), compared with \(.761/.659\) for its random-feature controls.

Appendix Table~\ref{tab:hallusae_causal_fp} reports the causal comparison. We
match 73 HalluSAE FP feature tests to our selected-feature budget within 42 dataset--model--layer--SAE-width.
HalluSAE produces substantially larger \( |\Delta m| \) than selected(ours) or random FP features, with differences of \(+1.606\)
\([1.327,1.902]\) and \(+1.524\) \([1.245,1.818]\), respectively.
However, its W2C rate is not detectably higher than either baseline, and its DLA-aligned margin interval includes zero. HalluSAE features are also much denser: their median activation frequency is \(.580\), compared with about \(.0013\) for selected and random FP features. The comparison therefore matches the feature-test budget, but not activation density. It means that HalluSAE identifies dense, artifact-prone directions with substantial logit influence, but not features that correct errors more reliably.

\section{Conclusion} \label{sec:conclusion}

We observed that across eight datasets, SAE activations contain predictive information about
correctness, confidence, and uncertainty. Our artifact-aware pipeline identifies
candidates whose Q/O interventions change correct-vs.-wrong margins more than
random-active controls and correct a larger fraction of initially wrong events.
Nevertheless, answer flips are rare, and effects are not stable across equivalent
prompt formats or answer orders. The selected features influence answer preference without
providing dependable error repair.

However, artifact entanglement remains the main issue. Although every final candidate passes
our prespecified gate, 26 of 55 non-oracle units retain an artifact or semantic
tag. HalluSAE is predictive and produces larger FP margin shifts, yet $95.5\%$
of its nonzero units fail our artifact threshold in every setting where they
are selected, without improving correction rates. Thus, neither prediction nor
large logit shifts establish clean error mechanisms. Our findings are consistent
with sparsity favoring low-dimensional surface cues while higher-level behavior
remains distributed across features~\cite{p72}.

\section*{Limitations} \label{sec:limitations}

Our results are limited to Gemma~3 1B and 4B at four
mid-to-late residual layers, and 16K/65K Gemma Scope SAE based on the available processing power. Larger models may show different behavior; however, the cost of analysis and intervention increases substantially.

Random-active controls are matched within activation runs rather than exactly
matched on activation frequency or edit magnitude. We therefore report
perturbation-adjusted analyses, although these do not include decoder norm, and
our bootstrap intervals condition on the fixed candidate shortlist. Our
HalluSAE comparison applies its feature-selection method to the same analyzed
layers rather than repeating its all-layer localization stage, and should
therefore be viewed as a matched-layer adaptation.

Finally, we focus on multiple-choice errors because the constrained answer space enables precise measurement of correct-versus-wrong logits and controlled feature interventions. Extending this analysis to free-form wrong answers is an important direction for future work.

\section*{Use of AI Assistance}
In preparing this work, we used AI-assisted tools for code completion with GitHub Copilot, and language editing, such as spell checking and stylistic revisions with Grammarly and ChatGPT. The authors take full responsibility for the scientific content, analyses, and conclusions presented in this work.

\bibliography{references}
\section*{Appendix}
\appendix

\setlength{\textfloatsep}{7pt plus 2pt minus 2pt}
\setlength{\floatsep}{6pt plus 2pt minus 2pt}
\setlength{\intextsep}{7pt plus 2pt minus 2pt}
\setlength{\dbltextfloatsep}{7pt plus 2pt minus 2pt}
\setlength{\dblfloatsep}{6pt plus 2pt minus 2pt}
\setlength{\abovecaptionskip}{3pt}
\setlength{\belowcaptionskip}{0pt}
\renewcommand{\topfraction}{0.95}
\renewcommand{\bottomfraction}{0.90}
\renewcommand{\textfraction}{0.05}
\renewcommand{\floatpagefraction}{0.80}
\renewcommand{\dbltopfraction}{0.95}
\renewcommand{\dblfloatpagefraction}{0.80}
\makeatletter
\setlength{\@fptop}{0pt}
\setlength{\@fpsep}{7pt plus 2pt minus 2pt}
\setlength{\@fpbot}{0pt plus 1fil}
\setlength{\@dblfptop}{0pt}
\setlength{\@dblfpsep}{7pt plus 2pt minus 2pt}
\setlength{\@dblfpbot}{0pt plus 1fil}
\makeatother

\section{Feature Ranking, Artifact Controls, and Clean-Feature Filters}
\label{app:feature_ranking}
\label{app:artifact_score}

Feature selection combines predictive evidence with artifact controls.
We first assign evidence weights to features identified by the
candidate-ranking probes. We then adjust and combine this evidence across
datasets, targets, response modes, and token positions. Finally, we apply
eligibility requirements and retain the highest-scoring feature units.
A feature unit denotes a particular feature within one model, layer,
and SAE width.

\paragraph{Initial candidate-ranking weights.}
For each candidate-ranking probe, we retain up to 100 features ordered by
the absolute magnitude of their fitted coefficients. Let \(j\) identify
a probe fitted for a particular dataset, model, layer, SAE width, target,
response mode, and token position. Feature \(f\), at coefficient rank
\(r_{f,j}\geq1\), contributes the initial evidence weight
\begin{equation}
    e^{(0)}_{f,j}
    =\frac{Q_j}{\log_2(r_{f,j}+1)},
\end{equation}
where \(Q_j\) is the validation quality of the probe producing that
coefficient. For binary targets, we use
\begin{equation}
    Q_{\mathrm{bin}}
    =\min\left\{1,
    \max\left[0,2(\mathrm{AUROC}_{\mathrm{val}}-0.5)\right]\right\}.
\end{equation}
For regression targets, we use
\begin{equation}
\begin{split}
    Q_{\mathrm{reg}}=\min\{1,\;&\max(0,R^2_{\mathrm{val}})\\
    &+0.25\max(0,|r_{\mathrm{val}}|-0.10)\},
\end{split}
\end{equation}
where \(r_{\mathrm{val}}\) is the Pearson correlation between predictions
and targets.

Binary probes receive zero weight at or below chance-level AUROC.
For regression probes, negative \(R^2\) values contribute zero, while
absolute prediction correlation above \(0.10\) contributes an additional
term weighted by \(0.25\). Both quality weights are capped at one\footnote{These are fixed ranking heuristics, not statistical significance measures.}. Evidence is retained only when \(Q_j>0\) and the feature
is active on at least 20 instances in the corresponding setting.

\paragraph{Artifact-risk score.}
We compute artifact controls separately for each dataset, model, layer,
SAE width, response mode, and token position. Token activations are
max-pooled within the position, and we consider features active on at
least 20 behavioral instances. Let \(A_f\) denote the resulting
instance-level activation vector, including zeros for inactive instances.

For a categorical variable \(g\), we measure association using
eta-squared:
\begin{equation}
    \eta^2(A_f,g)
    =
    \frac{\sum_c n_c(\bar A_{f,c}-\bar A_f)^2}
         {\sum_i(A_{i,f}-\bar A_f)^2},
\end{equation}
where \(n_c\) and \(\bar A_{f,c}\) are the size and mean activation of
category \(c\), and \(\bar A_f\) is the overall mean activation.
This quantity measures the fraction of activation variance explained
by category membership.

Let \(G_{\mathrm{ET}}\) group instances sharing the same original example
and prompt template, and let \(G_{\mathrm{EP}}\) group instances sharing
the same original example and answer permutation. Let \(R_G(A_f)\)
subtract the mean activation within each group \(G\).
The option-letter and prompt-template controls are
\begin{align}
    s_f^{\mathrm{letter}}
    &=\eta^2\!\left(R_{G_{\mathrm{ET}}}(A_f),y\right),\\
    s_f^{\mathrm{prompt}}
    &=\eta^2\!\left(R_{G_{\mathrm{EP}}}(A_f),
                   \mathrm{template}\right),
\end{align}
where \(y\) is the gold answer letter for each instance. The first
control measures whether activation tracks the gold letter across
answer-order permutations of the same example and template. The second
measures template sensitivity while holding the example and answer
ordering fixed.

Permutation instability measures the proportion of activation variance
not explained by example--template groups:
\begin{equation}
    s_f^{\mathrm{perm}}
    =1-\eta^2(A_f,G_{\mathrm{ET}}).
\end{equation}
Within each such group, instances differ in their answer ordering.

Topic/text dependence is the maximum available association with text
covariates:
\begin{equation}
\begin{split}
    s_f^{\mathrm{text}}
    =\max\{&\max_j|\mathrm{corr}(A_f,c_j)|,\\
           &\max_k\eta^2(A_f,g_k)\}.
\end{split}
\end{equation}
Here, \(\mathrm{corr}\) denotes Pearson correlation. Continuous covariates
\(c_j\) include text lengths, token counts, the number of answer choices,
answer-text TF-IDF similarities, and up to eight TF-IDF/SVD text
components. Categorical covariates \(g_k\) are the available nonconstant
question categories, present for CommonSenseQA and MMLU in our data.
Unavailable or constant covariates are omitted.

Each control is converted to its percentile \(q\) among the tested
features in the same setting. The percentile component of artifact
risk is
\begin{equation}
    a_f^{\mathrm{pct}}
    =\max\{q_f^{\mathrm{letter}},q_f^{\mathrm{prompt}},
           q_f^{\mathrm{perm}},q_f^{\mathrm{text}}\}.
\end{equation}
Thus, strong association with any one measured confound can produce a
high score. Please note that this is a relative score and not a probability of contamination by artifacts.

\paragraph{Eligibility and final selection.}
After aggregation, a feature unit is eligible only if its maximum
artifact risk is at most \(0.80\), its maximum activation frequency is
below \(0.95\), it is active on at least 100 instances in one contributing
setting, it has evidence from at least two datasets, and
\(\bar c_f\geq0.60\).
Activation support, density, dataset coverage, and coefficient-sign
consistency are separate eligibility requirements, not components of
the artifact-risk score.

\begin{figure*}[t]
\centering
\includegraphics[width=.98\textwidth]{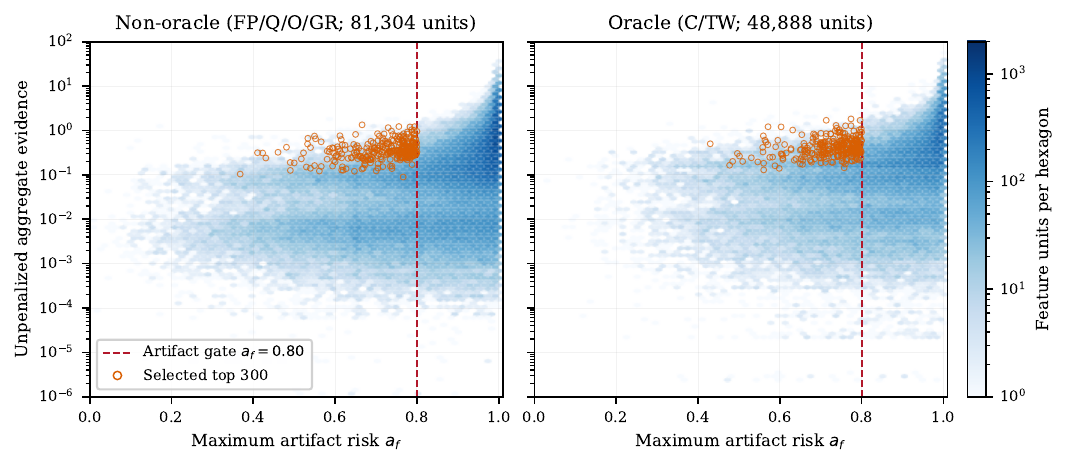}
\caption{Artifact risk versus unpenalized predictive evidence in the final
eight-dataset feature-selection pools. Each hexagon counts
model--layer--SAE-width--feature units; darker cells contain more units.
Evidence is the sum of validation-quality-weighted coefficient-rank scores
before artifact, density, overfit-gap, and position penalties. Orange circles
mark the selected top 300, and the dashed line marks the artifact-risk gate.
Because artifact risk is maximized and evidence is summed over contributing
candidate-ranking records, this descriptive visualization is not used as a correlation
test.}
\label{fig:artifact_score_vs_evidence}
\end{figure*}

\section{DLA-Based Candidate Selection}
\label{app:dla_thresholds}

We use DLA to identify shortlisted features whose decoder directions show both strong and consistent alignment with the answer contrast. Each combination of dataset, model, layer, SAE width, response mode, and activation position defines one setting. We use all DLA summaries from discovery and validation instances. We call a feature \emph{correct-directed} when its maximum correct-minus-top-wrong DLA is at least \(1.0\), its mean DLA in that direction exceeds \(0.02\), and the direction agrees across at least \(75\%\) of its matched settings. The feature must also rank among the top \(5\%\) of correct-directed DLA values in at least one setting, and its strongest correct-directed score must exceed its strongest top-wrong-directed score. The symmetric rule defines \emph{top-wrong-directed} features.

We retain large but context-dependent effects as \emph{mixed-direction} diagnostics. A feature gets this label when its maximum absolute DLA lies in the top \(10\%\) of the corresponding shortlist, ranks in the top \(5\%\) in both directions, and satisfies  neither directional rule. To prevent any model configuration from dominating, we retain at most five features per category and at most 25 distinct features per model--layer--SAE-width group. A directional feature also receives an artifact-control label when its artifact risk is at least \(0.70\), its description indicates a semantic or formatting pattern, or it is flagged as globally dense.

\section{Forced-Letter Behavioral Results}
\label{app:behavioral_results}

Table~\ref{tab:forced_letter_behavior} reports descriptive test behavior before SAE analysis. Metrics are averaged over prompt-template and answer-permutation instances. $N$ counts unique original examples and should not be interpreted as the number of independent rendered prompts. WHC is the fraction of all instances that are both wrong and assign at least $.70$ probability to the strongest wrong letter.

\begin{table}[t]
\centering
\scriptsize
\setlength{\tabcolsep}{2.5pt}
\begin{tabular}{lrrrrr}
\toprule
& & \multicolumn{2}{c}{1B} & \multicolumn{2}{c}{4B} \\
\cmidrule(lr){3-4}\cmidrule(lr){5-6}
Dataset & $N$ & Acc. & WHC & Acc. & WHC \\
\midrule
CommonSenseQA & 796 & .438 & .519 & .657 & .325 \\
TruthfulQA    & 175 & .209 & .735 & .501 & .485 \\
HaluEval QA   & 833 & .880 & .114 & .907 & .092 \\
Wiki-FACTOR   & 331 & .453 & .502 & .499 & .464 \\
Med-HALT      & 534 & .063 & .866 & .115 & .833 \\
MMLU          & 262 & .418 & .510 & .603 & .375 \\
SciQ          & 808 & .707 & .272 & .876 & .116 \\
FEVER         & 791 & .338 & .638 & .532 & .454 \\
\bottomrule
\end{tabular}
\caption{Forced-letter test accuracy and wrong high-confidence rate over
rendered prompt instances.}
\label{tab:forced_letter_behavior}
\end{table}

\section{Statistical Audit Details}
\label{app:statistical_screen}

Each model in the statistical audit is fit on discovery instances and evaluated on the test partition. A run corresponds to one dataset, model, layer, SAE width, response mode, and activation position. We use up to 1{,}000 features with at least 20 active instances. Binary models are standardized, class-balanced $L_1$-logistic regressions with $C=.1$; regression models are standardized ridge models with candidate regularization strengths $\{100\}$. Univariate analyses use Pearson and Spearman correlations with Benjamini--Hochberg correction, 100 bootstrap resamples for
binary mean-activation differences, and three-neighbor mutual information. Table~\ref{tab:stat_screen_target_definitions}
defines the six targets.

\begin{table}[t]
\centering
\small
\setlength{\tabcolsep}{4pt}
\begin{tabular}{llp{0.58\columnwidth}}
    \toprule
    Abbrev. & Type & Target \\
    \midrule
    WvC & Bin. & Wrong vs.\ correct answer \\
    LMvHM & Bin. & $m_i\leq0.5$ vs.\ $m_i\geq2.0$; intermediate instances are excluded \\
    WHC & Bin. & Wrong with $p_i(w_i)\geq0.70$ vs.\ correct with $p_i(y_i)\geq0.70$ or $m_i\geq2.0$ \\
    WLM & Bin. & Wrong vs.\ correct among instances with $m_i\leq0.5$ \\
    Margin & Reg. & Correct-vs.-top-wrong margin \\
    Entropy & Reg. & Entropy over answer letters \\
    \bottomrule
\end{tabular}
\caption{Statistical-audit target definitions.}
\label{tab:stat_screen_target_definitions}
\end{table}

Tables~\ref{tab:stat_screen_position_binary}
and~\ref{tab:stat_screen_position_regression} aggregate over datasets, models,
layers, widths, and response modes. Question-span models are weakest overall,
whereas GR and the oracle spans are stronger for several binary targets.
Regression is more heterogeneous: median $R^2$ is negative at Q for both
targets but positive at FP, O, and GR. Coverage is 192 runs each at FP, Q, O,
C, and TW, and 128 at GR because forced-letter runs have no generated
reasoning span.

\begin{table}[t]
\centering
\scriptsize
\setlength{\tabcolsep}{3pt}
\begin{tabular}{lcccc}
    \toprule
    Pos. & WvC & LMvHM & WHC & WLM \\
    \midrule
    FP & .715/.644 & .739/.711 & .739/.688 & .706/.805 \\
    Q  & .569/.522 & .558/.572 & .573/.553 & .567/.779 \\
    O  & .662/.591 & .720/.687 & .722/.662 & .638/.821 \\
    GR & .740/.660 & .749/.738 & .782/.732 & .784/.820 \\
    C  & .746/.652 & .836/.774 & .839/.760 & .645/.825 \\
    TW & .694/.617 & .801/.750 & .804/.735 & .595/.800 \\
    \bottomrule
\end{tabular}
\caption{Binary test performance by activation position. Each cell reports
mean AUROC/F1 over the available runs.}
\label{tab:stat_screen_position_binary}
\end{table}

\begin{table}[t]
\centering
\small
\begin{tabular}{lcc}
    \toprule
    Pos. & Margin $R^2$ / $r$ & Entropy $R^2$ / $r$ \\
    \midrule
    FP & .117/.357 & .285/.541 \\
    Q  & -.138/.102 & -.066/.176 \\
    O  & .059/.353 & .054/.302 \\
    GR & .104/.360 & .106/.373 \\
    C  & .239/.550 & -.020/.189 \\
    TW & .185/.480 & -.021/.196 \\
    \bottomrule
\end{tabular}
\caption{Regression test performance by activation position. Each cell reports
median $R^2$/Pearson $r$.}
\label{tab:stat_screen_position_regression}
\end{table}

\section{Statistical Audit by Dataset and Position}
\label{app:stat_screen_by_dataset_position}

Tables~\ref{tab:stat_screen_dataset_position_binary}
and~\ref{tab:stat_screen_dataset_position_regression} give the eight-dataset
breakdown. Each non-GR cell aggregates 24 runs (eight model--layer--SAE
configurations and three response modes); each GR cell aggregates 16 runs.
Negative $R^2$ indicates performance below a constant test-set predictor.

\begin{table}[t]
\centering
\scriptsize
\renewcommand{\arraystretch}{0.88}
\setlength{\tabcolsep}{4pt}
\begin{tabular}{llrrrr}
    \toprule
    Dataset & Pos. & WvC & LMvHM & WHC & WLM \\
    \midrule
    \multirow{6}{*}{CSQA}
     & FP & .652 & .688 & .670 & .648 \\
     & Q  & .539 & .531 & .530 & .555 \\
     & O  & .583 & .655 & .650 & .601 \\
     & GR & .672 & .727 & .751 & .719 \\
     & C  & .735 & .820 & .823 & .636 \\
     & TW & .646 & .781 & .787 & .556 \\
    \midrule
    \multirow{6}{*}{TruthfulQA}
     & FP & .722 & .717 & .730 & .710 \\
     & Q  & .622 & .602 & .629 & .589 \\
     & O  & .692 & .738 & .747 & .625 \\
     & GR & .802 & .740 & .832 & .795 \\
     & C  & .744 & .833 & .841 & .635 \\
     & TW & .716 & .800 & .818 & .591 \\
    \midrule
    \multirow{6}{*}{HaluEval}
     & FP & .865 & .902 & .905 & .759 \\
     & Q  & .582 & .568 & .583 & .549 \\
     & O  & .801 & .861 & .868 & .651 \\
     & GR & .905 & .919 & .932 & .895 \\
     & C  & .745 & .826 & .829 & .596 \\
     & TW & .783 & .848 & .854 & .631 \\
    \midrule
    \multirow{6}{*}{Wiki-FACTOR}
     & FP & .642 & .670 & .662 & .658 \\
     & Q  & .493 & .504 & .496 & .524 \\
     & O  & .558 & .621 & .620 & .561 \\
     & GR & .659 & .663 & .690 & .737 \\
     & C  & .694 & .807 & .809 & .615 \\
     & TW & .630 & .759 & .767 & .545 \\
    \midrule
    \multirow{6}{*}{Med-HALT}
     & FP & .809 & .851 & .857 & .748 \\
     & Q  & .591 & .579 & .605 & .561 \\
     & O  & .741 & .853 & .858 & .672 \\
     & GR & .826 & .826 & .842 & .808 \\
     & C  & .740 & .851 & .853 & .655 \\
     & TW & .605 & .732 & .720 & .561 \\
    \midrule
    \multirow{6}{*}{MMLU}
     & FP & .665 & .674 & .675 & .671 \\
     & Q  & .574 & .556 & .587 & .544 \\
     & O  & .613 & .610 & .622 & .600 \\
     & GR & .650 & .643 & .684 & .708 \\
     & C  & .685 & .785 & .793 & .593 \\
     & TW & .654 & .765 & .766 & .559 \\
    \midrule
    \multirow{6}{*}{SciQ}
     & FP & .735 & .771 & .760 & .686 \\
     & Q  & .555 & .544 & .558 & .585 \\
     & O  & .653 & .723 & .699 & .654 \\
     & GR & .745 & .811 & .795 & .794 \\
     & C  & .715 & .802 & .802 & .622 \\
     & TW & .657 & .771 & .768 & .581 \\
    \midrule
    \multirow{6}{*}{FEVER}
     & FP & .630 & .640 & .653 & .767 \\
     & Q  & .595 & .578 & .593 & .626 \\
     & O  & .655 & .698 & .709 & .738 \\
     & GR & .660 & .666 & .733 & .819 \\
     & C  & .910 & .962 & .965 & .806 \\
     & TW & .863 & .948 & .953 & .734 \\
    \bottomrule
\end{tabular}
\caption{Mean test AUROC for the four binary audit models by dataset and activation
position.}
\label{tab:stat_screen_dataset_position_binary}
\end{table}

\begin{table}[t]
\centering
\scriptsize
\renewcommand{\arraystretch}{0.88}
\setlength{\tabcolsep}{5pt}
\begin{tabular}{llrr}
    \toprule
    Dataset & Pos. & Margin & Entropy \\
    \midrule
    \multirow{6}{*}{CSQA}
     & FP & .062 & .369 \\
     & Q  & -.140 & -.017 \\
     & O  & .006 & .145 \\
     & GR & .105 & .174 \\
     & C  & .272 & -.019 \\
     & TW & .178 & .000 \\
    \midrule
    \multirow{6}{*}{TruthfulQA}
     & FP & .101 & .209 \\
     & Q  & -.390 & -.152 \\
     & O  & -.204 & -.102 \\
     & GR & .091 & .082 \\
     & C  & -.061 & -.141 \\
     & TW & .073 & -.107 \\
    \midrule
    \multirow{6}{*}{HaluEval}
     & FP & .504 & .124 \\
     & Q  & -.044 & -.048 \\
     & O  & .399 & .017 \\
     & GR & .471 & .054 \\
     & C  & .302 & -.021 \\
     & TW & .316 & -.016 \\
    \midrule
    \multirow{6}{*}{Wiki-FACTOR}
     & FP & .056 & .216 \\
     & Q  & -.222 & -.124 \\
     & O  & -.112 & -.050 \\
     & GR & .015 & .086 \\
     & C  & .172 & -.073 \\
     & TW & .094 & -.075 \\
    \midrule
    \multirow{6}{*}{Med-HALT}
     & FP & .506 & .350 \\
     & Q  & -.144 & -.050 \\
     & O  & .292 & .069 \\
     & GR & .169 & .126 \\
     & C  & .301 & .048 \\
     & TW & .023 & -.027 \\
    \midrule
    \multirow{6}{*}{MMLU}
     & FP & .095 & .257 \\
     & Q  & -.714 & -.584 \\
     & O  & -.140 & .038 \\
     & GR & .004 & .107 \\
     & C  & -.175 & -.095 \\
     & TW & .101 & -.059 \\
    \midrule
    \multirow{6}{*}{SciQ}
     & FP & .299 & .344 \\
     & Q  & -.155 & -.039 \\
     & O  & .130 & .092 \\
     & GR & .278 & .148 \\
     & C  & .233 & -.010 \\
     & TW & .173 & -.011 \\
    \midrule
    \multirow{6}{*}{FEVER}
     & FP & .068 & .363 \\
     & Q  & -.047 & .058 \\
     & O  & .116 & .287 \\
     & GR & .058 & .257 \\
     & C  & .752 & .161 \\
     & TW & .717 & .165 \\
    \bottomrule
\end{tabular}
\caption{Median test $R^2$ for margin and entropy regression by dataset and
activation position.}
\label{tab:stat_screen_dataset_position_regression}
\end{table}

\section{Prompt Templates}
\label{app:prompt_templates}

The unwrapped model input is the response-mode prefix, two newlines, and the corresponding rendered template body. Gemma's chat template then wraps the combined text as one user message. Table~\ref{tab:prompt_templates_exact} summarizes this construction; vertical gaps denote two newline characters.
Five labels are shown for readability, and the renderer substitutes the labels available for each dataset.

\begin{table*}[t!]
\centering
\scriptsize
\setlength{\tabcolsep}{4pt}
\renewcommand{\arraystretch}{1.10}
\textbf{Response-mode prefixes}\par\smallskip
\begin{tabularx}{\textwidth}{@{}l>{\raggedright\arraybackslash\ttfamily}X@{}}
\toprule
Mode & Exact prefix \\
\midrule
FL & You are a multiple-choice classifier. Reply with only the answer label. \\
RF & You are a careful multiple-choice assistant. Reason briefly about the choices, then put the final answer in the requested exact format on the last line. \\
LTR & You are a careful multiple-choice assistant. Start with the answer letter, then explain your reasoning briefly. \\
\bottomrule
\end{tabularx}

\vspace{0.6em}
\begin{tabularx}{\textwidth}{@{}l>{\raggedright\arraybackslash}X>{\raggedright\arraybackslash}X>{\raggedright\arraybackslash}X@{}}
\toprule
 & \textbf{T1: dotted labels} & \textbf{T2: parenthesized labels} & \textbf{T3: colon labels} \\
\midrule
Body
& {\ttfamily Question: <QUESTION>\par\smallskip
   A. <CHOICE\_A>\par B. <CHOICE\_B>\par C. <CHOICE\_C>\par
   D. <CHOICE\_D>\par E. <CHOICE\_E>\par\smallskip <MODE\_ENDING>}
& {\ttfamily Choose the best answer.\par\smallskip
   <QUESTION>\par\smallskip
   (A) <CHOICE\_A>\par (B) <CHOICE\_B>\par (C) <CHOICE\_C>\par
   (D) <CHOICE\_D>\par (E) <CHOICE\_E>\par\smallskip <MODE\_ENDING>}
& {\ttfamily <QUESTION>\par\smallskip Options:\par
   A: <CHOICE\_A>\par B: <CHOICE\_B>\par C: <CHOICE\_C>\par
   D: <CHOICE\_D>\par E: <CHOICE\_E>\par\smallskip <MODE\_ENDING>} \\
\midrule
FL ending
& {\ttfamily Answer:}
& {\ttfamily The correct option is}
& {\ttfamily Final answer:} \\
\addlinespace[0.35em]
RF ending
& {\ttfamily Reason briefly; final line: Final answer: <letter>\par
   Generation cue: Reasoning:}
& {\ttfamily Reason briefly; final line: Final answer: <letter>}
& {\ttfamily Reason briefly; final line: Final answer: <letter>} \\
\addlinespace[0.35em]
LTR ending
& {\ttfamily First write one answer letter (A/B/C/D/E), then explain briefly.\par\smallskip Answer:}
& {\ttfamily Give the letter first, then your reasoning.}
& {\ttfamily Answer letter first, then reasoning:} \\
\bottomrule
\end{tabularx}
\caption{Compact prompt construction for the three templates and three response
modes. Template 3, the colon-labeled options block, was used in all Phase~0
runs alongside Templates 1 and 2. For datasets with fewer choices, the option
lines and slash-separated label set are shortened accordingly.}
\label{tab:prompt_templates_exact}
\end{table*}

\paragraph{Answer parsing.}
For \emph{forced letter}, behavior is read from the next-token answer-letter distribution. For generated modes, decoding is deterministic. In
\emph{letter then reason}, the first matched answer letter is used. In
\emph{reason first}, explicit strings of the form \texttt{answer:
\textless letter\textgreater} or \texttt{final answer:
\textless letter\textgreater} are preferred; otherwise, the last matched answer letter is used. If no letter is parseable, we append
\texttt{Final answer:} and score one constrained valid label token.

\paragraph{Answer permutations.}
Answer-order permutations keep the same strings but reassign answer texts to
letters. 

\section{HalluSAE Predictive and Artifact Audit}
\label{app:hallusae_audit}

Table~\ref{tab:hallusae_predictive_artifact_audit} summarizes HalluSAE's
predictive performance, artifact-risk associations, and feature overlap with
our shortlists and intervention candidates.

\begin{table}[t]
\centering
\scriptsize
\setlength{\tabcolsep}{3.5pt}
\begin{tabularx}{\columnwidth}{@{}lXr@{}}
\toprule
Audit & Quantity & Result \\
\midrule
Predictive & Dataset--model--width groups & 32 \\
& HalluSAE test AUROC / F1 & .775 / .674 \\
& Random-control test AUROC / F1 & .761 / .659 \\
\midrule
Artifact & Nonzero feature--setting records & 3{,}020 \\
& Records above the $a_f=0.80$ gate & 2{,}987 (98.9\%) \\
& Unique nonzero feature units & 711 \\
& Units above the gate in every setting & 679 (95.5\%) \\
& Units below the gate in every setting & 9 (1.3\%) \\
& Units that cross the gate across settings & 23 (3.2\%) \\
& Units with a dedicated artifact flag & 578 (81.3\%) \\
\midrule
Overlap & Our top-300 / Phase~2 units & 0 / 0 \\
\bottomrule
\end{tabularx}
\caption{Adapted HalluSAE predictive and artifact audit. Predictive metrics
are means over the 32 groups; random-control metrics average ten repetitions
per group. A unit is a model--layer--SAE-width--feature combination. Artifact
categories refer to the forced-letter FP settings in which a unit has a
nonzero classifier coefficient.}
\label{tab:hallusae_predictive_artifact_audit}
\end{table}

\section{Detailed Causal Intervention Results}
\label{app:causal_details}

Table~\ref{tab:phase2_pair_level} distinguishes intervention-event counts from
unique feature--instance pairs. Table~\ref{tab:hallusae_causal_fp} compares
HalluSAE with selected and random-active FP features, while
Table~\ref{tab:phase2_uncertainty} reports pooled Q/O selected--random
differences and oracle C/TW outcomes with bootstrap intervals. Each sampled
feature--instance pair is evaluated under zero replacement, mean replacement,
and steering by one, two, and four feature standard deviations. Random-active
controls are sampled within each run; C and TW have no random-active
counterpart.

\begin{table*}[t!]
\centering
\scriptsize
\setlength{\tabcolsep}{4pt}
\begin{tabular}{llrrrrrrrr}
\toprule
Set & Pos. & $U_r/U_f$ & Pairs & W2C e/p & C2W e/p & W2W e/p
& W2C\% & C2W\% & $\overline{|\Delta m|}$ \\
\midrule
Selected & FP & 73/20   & 500   & 2/1  & 9/3  & 9/3  & .159 & .723 & .154 \\
Random   & FP & 320/311 & 1{,}990 & 18/10 & 22/9 & 19/9 & .347 & .462 & .237 \\
HalluSAE & FP & 73/27   & 730   & 5/3  & 9/9  & 3/3  & .332 & .420 & 1.760 \\
\midrule
Selected & Q  & 425/55  & 4{,}222 & 16/6 & 17/6 & 8/6  & .170 & .145 & .158 \\
Random   & Q  & 320/319 & 3{,}030 & 1/1  & 9/3  & 10/6 & .013 & .119 & .135 \\
Selected & O  & 415/55  & 4{,}034 & 19/8 & 19/10 & 23/12 & .209 & .172 & .216 \\
Random   & O  & 320/318 & 2{,}679 & 7/7  & 4/2  & 3/2  & .111 & .056 & .159 \\
\midrule
Oracle   & C  & 145/22  & 1{,}361 & 6/2  & 9/4  & 4/2  & .211 & .227 & .132 \\
Oracle   & TW & 160/22  & 1{,}558 & 0/0  & 5/3  & 0/0  & .000 & .107 & .135 \\
\bottomrule
\end{tabular}
\caption{Event- and pair-level intervention outcomes. $U_r/U_f$ gives
run-specific and global model--layer--width--feature units. ``e/p'' gives
intervention events and unique feature--instance pairs. W2C\% is conditional
on an initially wrong baseline answer; C2W\% is conditional on an initially
correct answer.}
\label{tab:phase2_pair_level}
\end{table*}

\begin{table*}[t!]
\centering
\scriptsize
\setlength{\tabcolsep}{4pt}
\begin{tabular}{lccccc}
\toprule
Metric & Selected FP & Random FP & HalluSAE FP
& H--Selected $[95\%\ \mathrm{CI}]$ & H--Random $[95\%\ \mathrm{CI}]$ \\
\midrule
Run-specific feature tests & 73 & 320 & 73 & -- & -- \\
Feature--instance pairs & 500 & 1{,}990 & 730 & -- & -- \\
Median activation frequency & .00133 & .00131 & .580 & -- & -- \\
\midrule
W2C rate (\%) & .159 & .347 & .332
& +.173 $[-.294,.745]$ & -.015 $[-.458,.557]$ \\
C2W rate (\%) & .723 & .462 & .420
& -.303 $[-1.489,.566]$ & -.043 $[-.483,.412]$ \\
$\overline{|\Delta m|}$ & .154 & .237 & 1.760
& +1.606 $[1.327,1.902]$ & +1.524 $[1.245,1.818]$ \\
$\overline{|\Delta p_y|}$ & .00305 & .00392 & .00517
& +.00212 $[-.00110,.00567]$ & +.00125 $[-.00169,.00462]$ \\
\bottomrule
\end{tabular}
\caption{Matched-budget HalluSAE FP comparison. Outcome intervals use 10{,}000
bootstrap resamples of model--layer--width--feature units. HalluSAE and
selected FP have the same feature-test budget within 42 strata; activation
density and the number of active feature--instance pairs are not matched.}
\label{tab:hallusae_causal_fp}
\end{table*}

\begin{table*}[t!]
\centering
\scriptsize
\setlength{\tabcolsep}{5pt}
\resizebox{0.98\textwidth}{!}{
\begin{tabular}{lcccc}
\toprule
Metric & Selected Q/O & Random Q/O & Selected--random & Oracle C/TW \\
\midrule
W2C rate (\%)
& .189 [.090, .308] & .058 [.021, .101] & +.132 [+.024, +.257] & .101 [.000, .307] \\
C2W rate (\%)
& .158 [.074, .253] & .089 [.014, .179] & +.069 [-.055, +.191] & .162 [.024, .330] \\
$\overline{|\Delta m|}$
& .187 [.164, .212] & .146 [.136, .160] & +.041 [+.014, +.067] & .134 [.114, .154] \\
$\overline{|\Delta p_y|}$
& .00152 [.00122, .00187] & .00133 [.00114, .00152]
& +.00019 [-.00018, +.00059] & .00105 [.00085, .00130] \\
\bottomrule
\end{tabular}
}
\caption{Point estimates and 95\% percentile intervals from 5{,}000 bootstrap
resamples of model--layer--width--feature units. The intervals describe
variation across tested feature units; they are not row-level tests or
randomization-based $p$-values.}
\label{tab:phase2_uncertainty}
\end{table*}

\begin{figure}[tbp]
\centering
\includegraphics[width=.95\columnwidth]{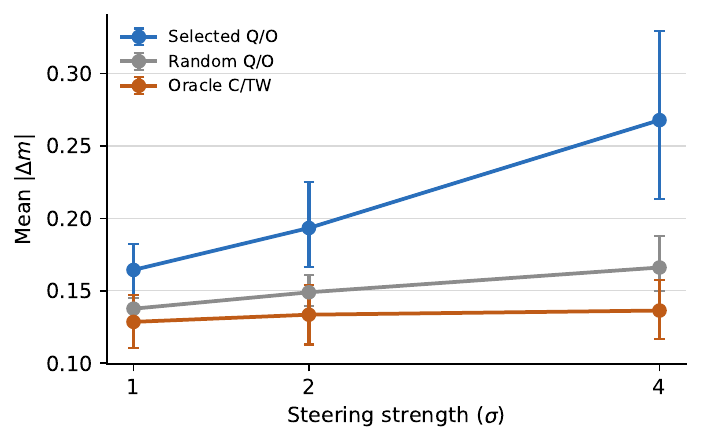}
\caption{Mean absolute margin shift under upward steering. Error bars are 95\%
feature-unit bootstrap intervals over matched pairs with $+1\sigma$,
$+2\sigma$, and $+4\sigma$ results.}
\label{fig:phase2_dose_response}
\end{figure}

\paragraph{Steering-strength response and perturbation adjustment.}
Figure~\ref{fig:phase2_dose_response} uses only feature--instance pairs observed
at all three steering strengths. Its intervals resample
model--layer--width--feature units. Selected Q/O has the clearest aggregate
strength response, but only $51.8\%$ of its 8{,}256 pairs increase monotonically;
the corresponding fractions are $46.6\%$ for random Q/O and $48.6\%$ for
oracle C/TW.

\paragraph{Adjustment for intervention size and context.}
Table~\ref{tab:phase2_adjusted_effects} compares the unadjusted and adjusted
selected--random differences in mean $|\Delta m|$.
Because the magnitude of an intervention depends partly on how far the feature
is moved and how many tokens are edited, we perform an intervention-size
adjustment. For each event, we measure the \(L_2\) displacement between the
original and target feature activations across edited tokens, normalized by the
feature's activation standard deviation. A feature-equal regression then
compares selected and random features while controlling for this displacement,
span length, intervention variant, baseline margin and correctness, activation
frequency, token position, dataset, and model configuration. The resulting
coefficient estimates the selected--random difference in \( |\Delta m| \) for
interventions with comparable measured characteristics.

\begin{table}[tbp]
\centering
\scriptsize
\setlength{\tabcolsep}{2.8pt}
\resizebox{\columnwidth}{!}{%
\begin{tabular}{lcccc}
\toprule
Scope & Tokens S/R & Std. edit S/R & Raw diff. & Adjusted diff. [95\% CI] \\
\midrule
Q/O & 33/34 & 11.49/14.19 & .041 & .021 $[-.006,.048]$ \\
Q   & 32/35 & 11.62/14.39 & .023 & .008 $[-.001,.017]$ \\
O   & 33/33 & 11.38/13.94 & .058 & .049 $[.014,.084]$ \\
\bottomrule
\end{tabular}%
}
\caption{Selected--random difference in mean $|\Delta m|$ before and after
perturbation adjustment. Tokens and standardized latent-edit magnitude are
medians for selected/random events. The feature-equal regression additionally
adjusts for baseline margin and correctness, activation frequency, operation,
dataset, and model configuration; intervals are clustered by feature and
original question. Standardized latent magnitude excludes decoder norm.}
\label{tab:phase2_adjusted_effects}
\end{table}

\section{Model Size and Budget}

For running the experiments, we used 20 Nvidia V100 SXM2 GPUs. We used approximately 500 GPU hours for the experiments and analysis.

\end{document}